\RequirePackage{amsmath}
\RequirePackage{graphicx}
\RequirePackage{amssymb}
\RequirePackage{interval}
\RequirePackage{algorithm}
\RequirePackage{algpseudocode}

\RequirePackage{color}

\documentclass[runningheads]{llncs}

\begin{document}

\title{Photometric Transformer Networks and \\ Label Adjustment for Breast Density Prediction}
\titlerunning{Breast Density Prediction in Mammograms}

\author{Jaehwan Lee \and Donggeon Yoo \and Jung Yin Huh \and Hyo-Eun Kim}
\authorrunning{J. Lee et al.}
%
\institute{Lunit Inc., Seoul, South Korea}
\maketitle              

\begin{abstract}

Grading breast density is highly sensitive to normalization settings of digital mammogram as the density is tightly correlated with the distribution of pixel intensity. 
Also, the grade varies with readers due to uncertain grading criteria.
These issues are inherent in the density assessment of digital mammography. 
They are problematic when designing a computer-aided prediction model for breast density and become worse if the data comes from multiple sites.
In this paper, we proposed two novel deep learning techniques for breast density prediction: 1) \emph{photometric transformation} which adaptively normalizes the input mammograms, and 2) \emph{label distillation} which adjusts the label by using its output prediction.
The \emph{photometric transformer network} predicts optimal parameters for photometric transformation on the fly, learned jointly with the main prediction network.
The \emph{label distillation}, a type of pseudo-label techniques, is intended to mitigate the grading variation.
We experimentally showed that the proposed methods are beneficial in terms of breast density prediction, resulting in significant performance improvement compared to various previous approaches.

\end{abstract}

\section{Introduction}

Breasts can be categorized as \emph{dense} or \emph{fatty} by the portion of parenchyma in the breasts.
A fatty breast indicates that the breast is mostly composed of fat tissue, whereas a dense breast has more dense tissue that shows dense parenchymal patterns on mammograms.
Readers should be more careful when dealing with mammograms with dense parenchymal pattern since suspicious malignant lesions can be hidden, resulting to a false-negative~\cite{Kerlikowske1996}.
Also, it has been reported that a dense breast has a higher risk of breast cancer than average~\cite{Boyd2007}.
For this reason, BI-RADS~\cite{Sickles2013}, which is a standard protocol for breast imaging, guides the interpreting readers to report density category as an essential field of case reports form(CRF).
In BI-RADS taxonomy, breast density is categorized into four grades: \emph{a}, \emph{b}, \emph{c}, \emph{d}, meaning ``almost entirely fatty'', ``scattered areas of fibro-glandular tissue'', ``heterogeneously dense'', and ``extremely dense'', respectively.

Based on the collected mammograms and their density categories in CRFs, it is straight-forward to regard a density prediction task as classification.
However, breast density prediction is not a typical classification task.
The BI-RADS criteria for breast density are 1) the portion of parenchyma within a breast, which is discretization of the continual score, and 2) specific dense parenchyma pattern in part of the image, determined by the reader.
Thus, the density labels in a training dataset will have inter-readers biases.

Intensity normalization of mammograms is an important factor when grading the breast density, since the mammographic parenchymal pattern is highly correlated with the pixel intensity.
However, intensity distribution of the parenchyma and the fat tissue varies according to different vendors of imaging devices as well as different hospitals.
To compensate these variations, readers often manually adjust the contrast of each mammogram to determine the grade properly.

In this paper, we propose two methods that tackle the problems caused by the normalization and inter-reader grading variance.
The first method is a learnable normalization module, called photometric transformer network (PTN), that predicts normalization parameters of input mammogram.
It is seamless to main prediction network so that optimal normalization and density grade can be learned jointly.
The second one is a label distillation method, which is a type of pseudo-label technique, taking the grading variation into consideration.

Our test shows that proposed two methods help to improve performance, especially in multi-site configurations.
Our final model outperforms other public-available previous models in a test set with neutral configurations.

\section{Related works}
With the drastic advance of deep learning, breast density prediction based on deep neural networks has also been introduced recently.
\cite{Kallenberg2016} applied the unsupervised feature learning based on auto-encoder to predict the breast density.
\cite{Lehman2018,Mohamed2018,Wu2018} employed convolutional neural networks (CNNs) that is learned with a cross-entropy loss for breast density prediction.
Motivated by these approaches, we also cast the breast density prediction as a CNN-based classification task, but address the two practical problems caused by multi-site configuration.

From the perspective of dynamic estimation of the parameters which are appropriate for a target task, our PTN is similar to the spatial transformer network \cite{NIPS2015_5854}.
Spatial transformer network predicts appropriate geometric transformation parameters, while our PTN tries to find a set of photometric transformation parameters that is optimal for breast density prediction.

The proposed label distillation is motivated by pseudo-labeling techniques, devised especially for handling label noise \cite{Li2017,Tanaka2018}.
In \cite{Li2017}, an auxiliary network trained with small clean examples were used to predict pseudo-labels, in addition to the main network trained with large examples with given pseudo-labels.
Similarly in \cite{Tanaka2018}, a sub-network jointly optimized with a main network tries to find appropriate pseudo-labels.
Our approach is distinct from~\cite{Li2017,Tanaka2018}, in that pseudo-labels are given to only selected samples and applied in iterative ways to prevent distillation of model bias.

\section{Methods}

A density estimator $f$ is a neural network that predicts breast density $y \in \{a, b, c, d\}$ from an input mammogram $x \in \mathbb{R}^{H \times W}$.
The input $x$ is normalized by the PTN denoted by $f_n$, and the classifier $f_c$ estimates density $\hat{y}$ from the normalized input as
\begin{equation} \label{eq:evaluation}
    \hat{y} = f_c(f_n(x; \theta_n); \theta_c),
\end{equation}
where $f_n$ and $f_c$ are parameterized by $\theta_n$ and $\theta_c$, respectively.
Our goal is to learn parameters $\theta = \theta_n \cup \theta_c$ with our dataset $D = \{(x_i, y_i) \, | \, i=1,\cdots,N\}$.
\begin{equation} \label{eq:training}
    \theta^* = \arg\min_{\theta}{\frac{1}{N}\sum_{(x,y)\in D}\mathcal{L}(\hat{y}, y)}
\end{equation}
where  $\mathcal{L}$ is the loss function.
To successfully estimates $\theta$, we propose the photometric transformer module $f_n$ in Section~\ref{ssec:ptn}, and a distillation method to handle label grading variance problem in Section~\ref{ssec:psudo_labeling}.


\subsection{Photometric transformer networks} \label{ssec:ptn}

\begin{figure}[t]
    \begin{center}
        \includegraphics[width=0.74\textwidth]{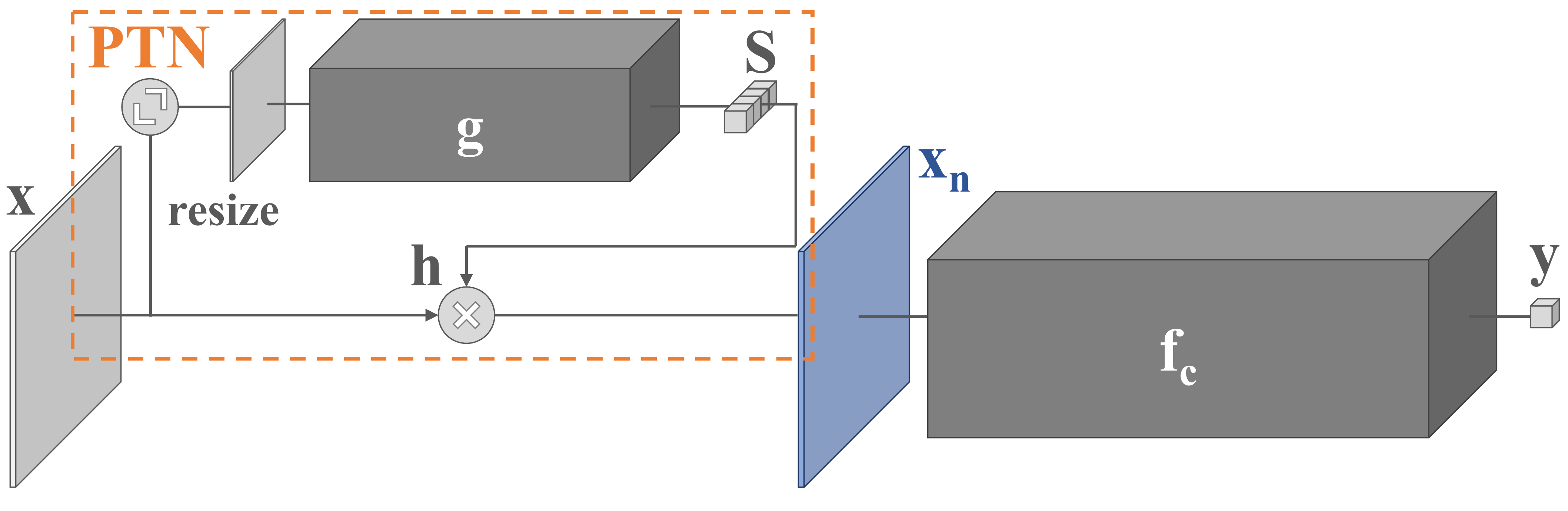}
        \includegraphics[width=0.24\textwidth]{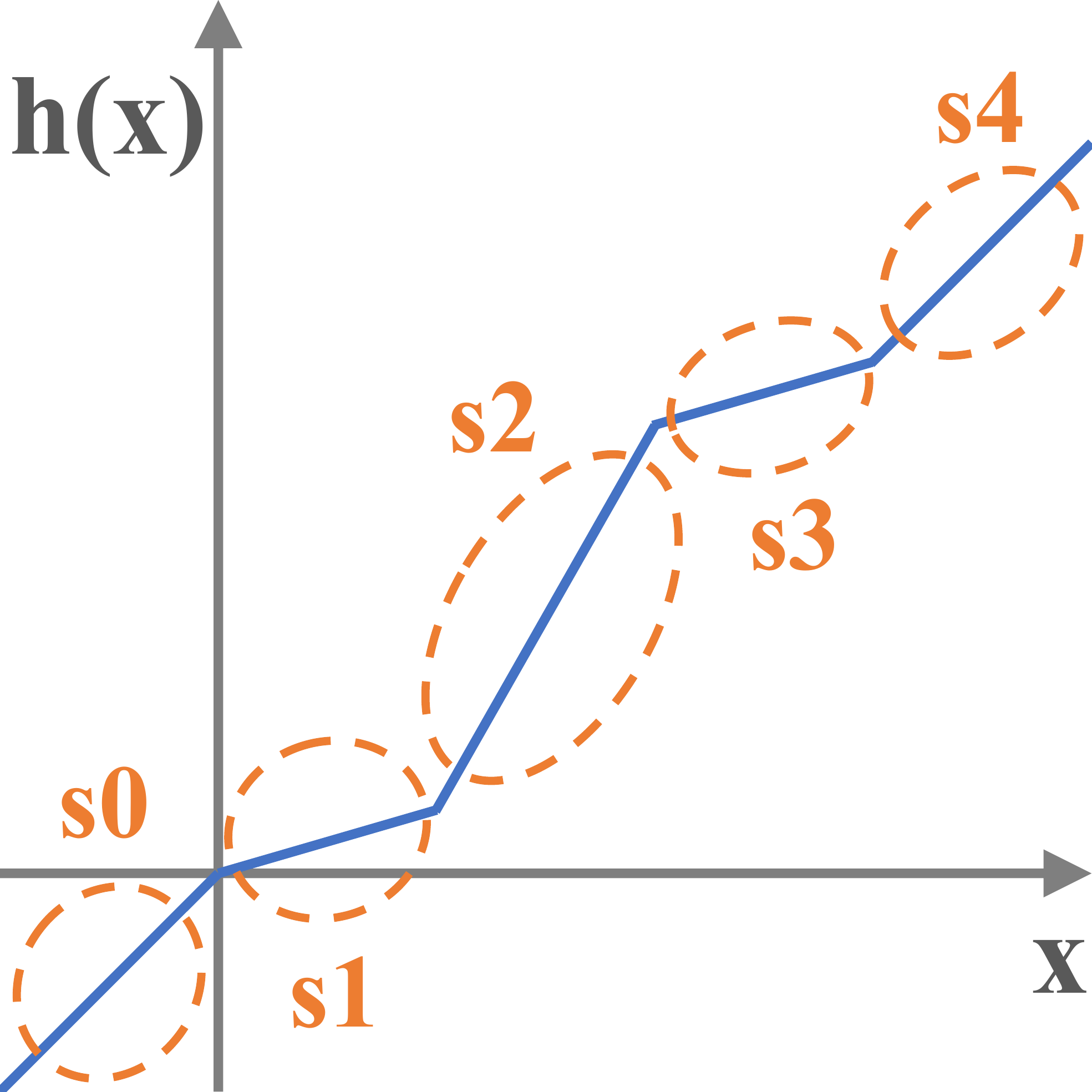}
    \end{center}
    \caption{
            (left) An overview of the model architecture.
            (right) Proposed function $h$.
        }
    \label{fig:architecture}
\end{figure}

The $f_n$ normalizes an input $x$ by a function $h$.
The function $h$ is determined by a parameter set $S$, and the parameter set $S$ is predicted by a CNN $g$ from the input $x$. For a pixel intensity $x(i,j)$ at a location $(i,j)$, it can be expressed as
\begin{equation} \label{eq:ptn}
x_n(i, j) = h(x(i, j), S)\quad\text{where}\quad S=g(x; \theta_{n}),
\end{equation}
and is illustrated in the left of Figure~\ref{fig:architecture}.

\intervalconfig{soft open fences}
We introduce the function $h$ that works well in breast density prediction.
Let us assume the intensity range of interests is $\interval[open right]{u}{v}$\footnotemark{}.
\footnotetext{Generally, it is determined by window center \& width value in standard DICOM.}
We split the range into $K$ sub-intervals, giving $T_k = \interval[open right]{u + t(k - 1)}{u + tk}$ where $t = (v-u)/K$ and $k=1,\cdots,K$.
Then, the function $h$ is defined as
\begin{equation} \label{eq:sampler}
    h(x(i, j), S) = 
        \begin{cases}
            u + s_0(x(i, j) - u) &\quad \text {if} \ x(i, j) \in \interval[open]{-\infty}{u} \\
            u + \sum_{l=1}^{k-1}{s_l t} + s_k (x(i, j) - \min(T_k)) &\quad \text {if} \ x(i, j) \in T_k \\
            u + \sum_{l=1}^{k}{s_l t} + s_{K+1}(x(i, j) - v) &\quad \text {if} \ x(i, j) \in \interval[open right]{v}{\infty}
        \end{cases}
\end{equation}
where $S$ is $\{s_0,\cdots,s_{K+1}\}$, and $\min(T_k)$ is a minimum value of an interval $T_k$.
The right of Figure~\ref{fig:architecture} is illustrating this function $h$.
Each component of $S$ can be interpreted as a slope of the corresponding line segment.

The function $h$ is continuous but can fluctuate if a part of $S$ is negative.
To make $h$ be an increasing function, we add a hinge regularization term to the cross entropy loss $\mathcal{L}_\text{CE}$.
The loss function in Equation~(\ref{eq:training}) is finally defined as
\begin{equation}
    \label{eq:loss}
    \mathcal{L}(\hat{y},y) = \mathcal{L}_\text{CE}(\hat{y},y) + \lambda\cdot\sum_{i=0}^{k+1}{\min(-s_i, -\epsilon) + \epsilon},
\end{equation}
where $\epsilon$ is a small positive constant and $\lambda$ is a scaling constant.
We have empirically found that adding this regularization term yields better performance.

\subsection{Label distillation} \label{ssec:psudo_labeling}
\begin{algorithm}[t]
    \caption{Label distillation}
    \label{alg:label_distillation}
    \begin{algorithmic}
        \State $\theta := \arg\min_{\theta}{\frac{1}{N}\sum_{(x,y)\in D}\mathcal{L}(f(x;\theta), y)}$
        \State Split $D$ into $D_s$ and $D_r$
        \While{not converged}
            \State $\theta$ := $\arg\min_{\theta}{\frac{1}{N}\sum_{(x,y)\in D_s}\mathcal{L}(f(x;\theta), y)}$
            \For{$(x,y)\in D_r$}
                \If{KLD($y,f(x;\theta)$) is top r\%}
                    \State $y$ := $\alpha y+(1-\alpha)\cdot f(x;\theta)$
                \EndIf
            \EndFor
            \State $D_\text{train}$ := $D_s\cup D_r$
            \State $\theta$ := $\arg\min_{\theta}{\frac{1}{N}\sum_{(x,y)\in D_\text{train}}\mathcal{L}(f(x;\theta), y)}$
        \EndWhile 
        \State \Return $\theta$
    \end{algorithmic}
\end{algorithm}
We propose a two-stage learning approach to alleviate the inter-readers grading variance problem.
For this, we split the dataset $D$ into a small set $D_s$ labeled by a single reader and the rest $D_r$.
$D_s$ is free from inter-reader variance, while $D_r$ is not.
We use a model trained with $D_s$ to evaluate $D_r$ and obtain pseudo-label $\{\hat{y}|x\in D_r\}$ from the model.
The model is pre-trained with $D$ and fine-tuned with $D_s$ since it is a small set.

In order to propagate the grading criterion of the single reader to the rest of data $D_r$, we update the labels in $D_r$ by taking $\{\hat{y}|x\in D_r\}$ into consideration as follows.
First, we measure a Kullback-Leibler divergence $\text{KLD}(y,\hat{y})$ between a one-hot encoded label $y$ and prediction $\hat{y}$ for each sample $x\in D_r$, and select top $\gamma$-percent samples of $\text{KLD}(y,\hat{y})$. The use of KLD is intended to select the most suspicious samples whose labels $y$ and strong predictions $\hat{y}$ do not match. After that, for each of the selected samples, we update $y$ with $\hat{y}$ by blending operation as
\begin{equation}
y:=\alpha y+(1-\alpha)\cdot\hat{y},
\end{equation}
where $\alpha$ is a constant blending factor. We then continue training with $D_s$ and the updated $D_r$. This procedure from $D_s$ training to $D_r$ training is repeated until the performance converges. Algorithm~\ref{alg:label_distillation} is concretely describing our label distillation.

\section{Evaluation}

\subsection{Experimental setup}

\subsubsection{Datasets}
\begin{table}[t]
    \caption{Dataset configurations}
    \label{tab:dataset}
    \begin{center}
        \begin{tabular}{l|r|r|r|r|r}
            \hline \hline
            Datasets$\backslash$Grades& $a$ & $b$ & $c$ & $d$ & Total \\ \hline
            Training set $D_r$ & 1,395 & 6,905 & 33,282 & 4,773 & 46,355 \\
            Training set $D_s$ & 72 & 391 & 428 & 255 & 1,146 \\
            Validation set & 78 & 373 & 421 & 275 & 1,147 \\
            Test set & 9 & 280 & 455 & 242 & 986 \\
            External test set & 852 & 3,130 & 3,634 & 590 & 8,206 \\
            \hline \hline
        \end{tabular}
    \end{center}
\end{table}
We have collected 48,648 cases of Asian women from 5 separate hospitals from South Korea.
Each case comprises four mammograms with different views of a left CC, a left MLO, a right CC, and a right MLO.
We also select approximately 5\% samples to refine labels by a single reader, i.e., a radiologist who is a breast specialist.
Half of the 5\% samples are used for $D_s$, and the rest for a validation set.
As an in-house test set, we have collected 986 cases from another institution from South Korea.
The same radiologist has labeled this test set.
To fairly compare ours with other method, we have collected another test set, which comprises 8,206 cases, from a large hospital in the US.
We have extracted the density grade for each case from CRF field, and use it as a label. Table~\ref{tab:dataset} summarizes datasets.

\subsubsection{Baseline}
For classifier $f_c$, we adopt ResNet-18 and make it produce a 4-dimensional softmax output.
We used SGD optimizer and the learning rate is set to $0.1$ in training.
The model takes a single mammogram as input, and four predictions from four views are averaged to a case-level prediction.
We decode mammograms by the window center and width embedded in DICOMs.



\subsubsection{Metrics}


We use the 4-way classification accuracy, as it has been the common metric for previous works.
However, breast density prediction is not a typical classification task.
Since the density grade is discretization of a continual density score, there exist relations between grades.
The grade $a$ is closer to $b$ than $c$ and $d$.
For this reason, the traditional accuracy metric is sensitive to inter-reader grading variation.
To take these relations into account, we propose a new metric, called density-AUC, which is more suitable for this problem.
The density-AUC (dAUC) is measured by averaging three AUC scores over: [$a$ vs. $b$, $c$, $d$], [$a$, $b$ vs. $c$, $d$], and [$a$, $b$, $c$ vs. $d$].
For instance, when we measure an AUC score of [$a$, $b$ vs. $c$, $d$], a sample score is defined as $\hat{y_a}$+$\hat{y_b}$ where $\hat{y_a}$ and $\hat{y_b}$ are the two elements of softmax output $\hat{y}$.


\subsection{Results and analysis}
\label{ssec:analysis}

\begin{figure}[t]
    \begin{center}
    \begin{tabular}{cccc}
        \includegraphics[height=0.25\textwidth]{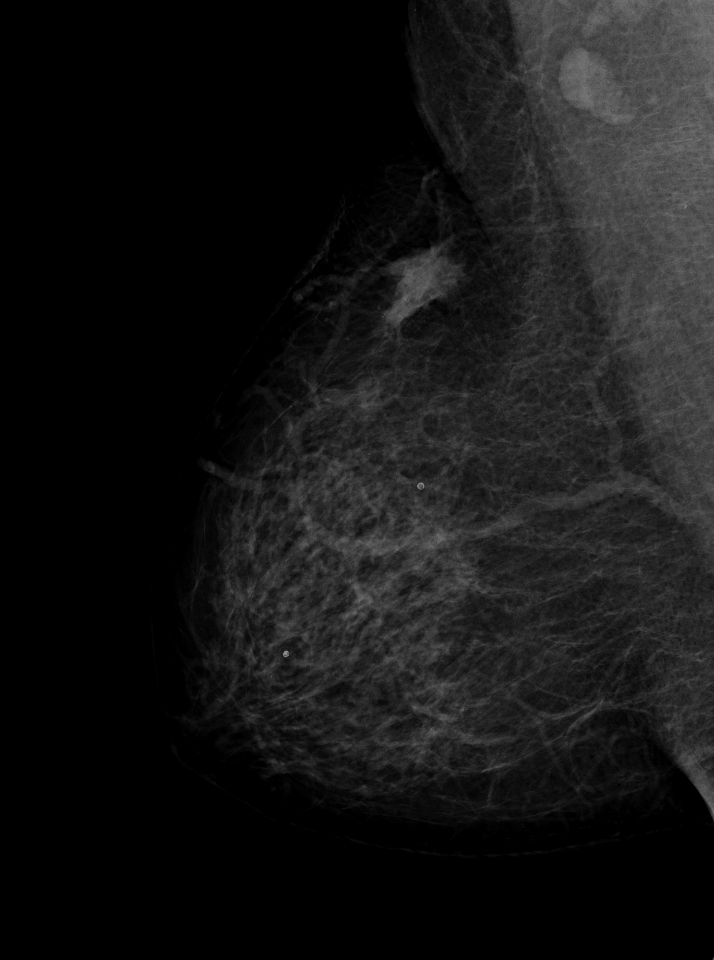}&
        \includegraphics[height=0.25\textwidth]{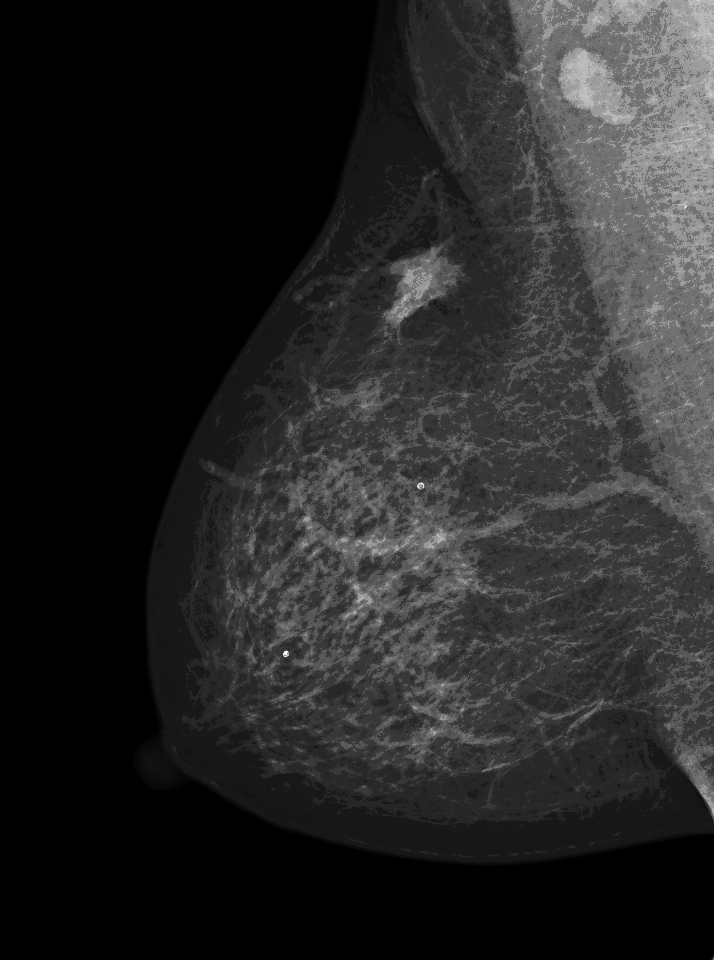}&
        \includegraphics[height=0.25\textwidth]{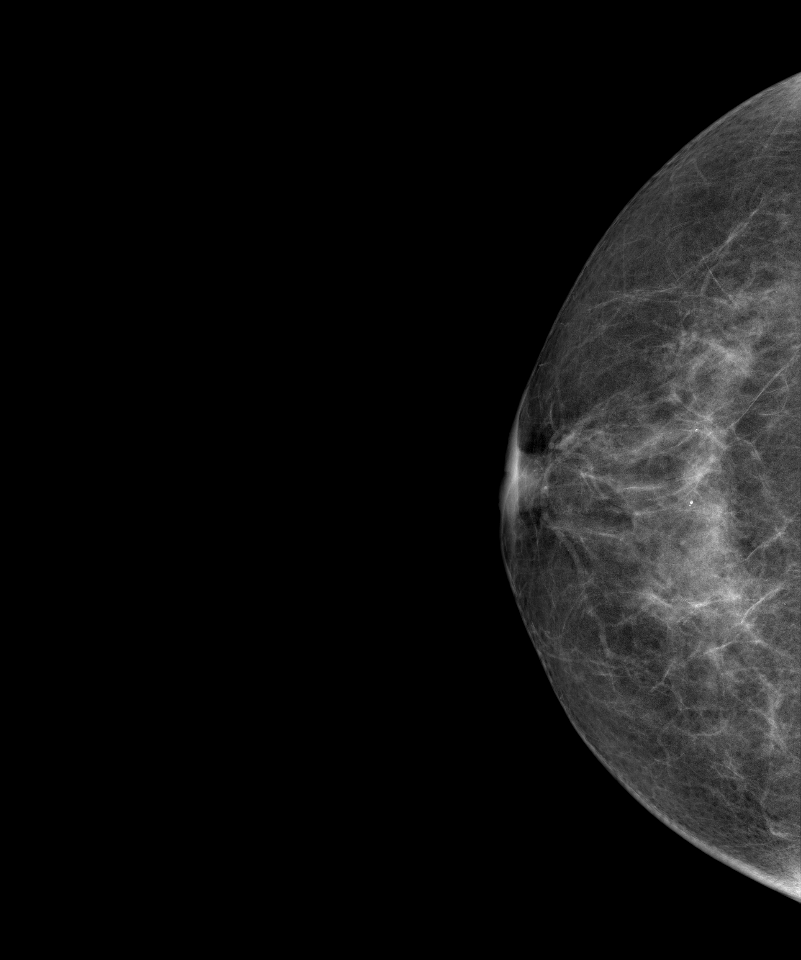}&
        \includegraphics[height=0.25\textwidth]{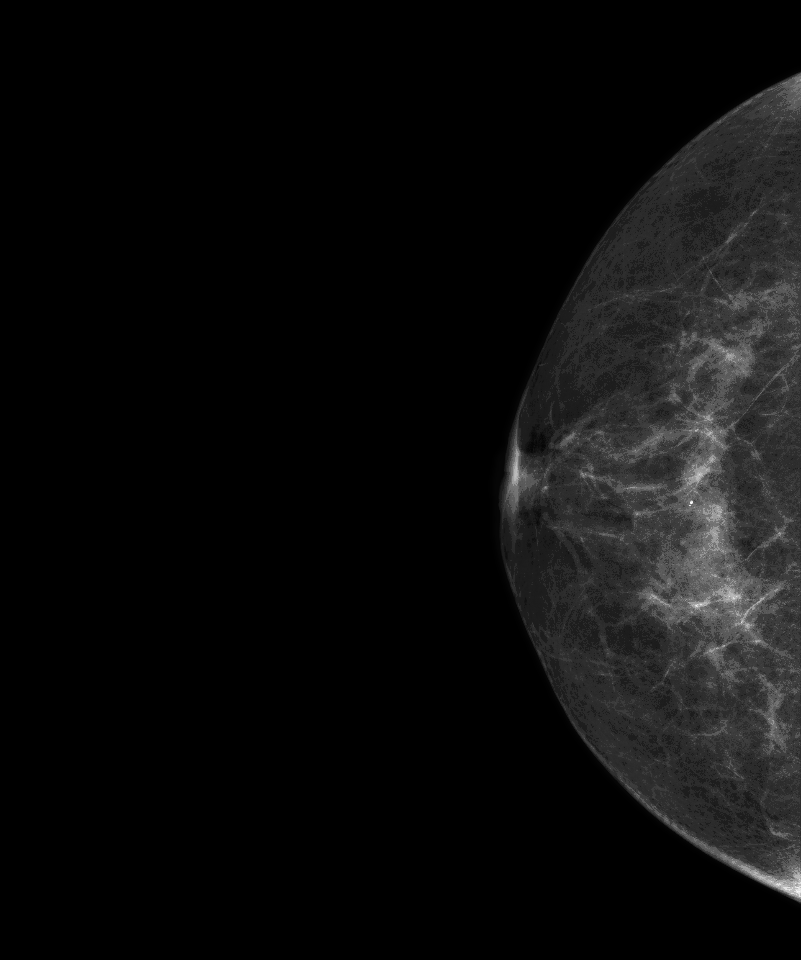}\\
        (1) Input & (2) Normalized & (3) Input & (4) Normalized
    \end{tabular}
    \end{center}
    \caption{Original mammograms (1), (3) with the density \emph{a} and \emph{d} are normalized to (2), (4) by PTN.
    For both samples, labels are corrected to \emph{b}.}
    \label{fig:qualitative_results}
\end{figure}

\begin{table}[t]
    \caption{Breast density estimation performance comparison between methods. The mean and standard deviation of 5 trials are reported.}
    \label{tab:algo}
    \begin{center}
        \begin{tabular}{l|c|c|c|c}
            \hline \hline
            & \multicolumn{2}{c|}{Validation} & \multicolumn{2}{c}{Test} \\ \cline{2-5}
            Methods & Accuracy & dAUC & Accuracy & dAUC \\ \hline
            Baseline & .7015(.0179) & .9595(.0153) & .5452(.1078) & .9204(.0207) \\
            CLAHE~\cite{Pizer1987} & .7374(.0291) & .9654(.0154) & .7163(.0341) & .9357(.0128) \\
            PTN & .7479(.0229) & .9755(.0013) & .7509(.0103) & .9518(.0079) \\
            \hline
            PTN\footnotemark{} & .7512(.0109) & .9757(.0014) & .7431(.0046) & .9470(.0045) \\
            PTN + hard labeling & .7367(.0150) & .9715(.0026) & .7671(.0039) & .9392(.0155) \\
            PTN + \cite{Tanaka2018} & .7428(.0126) & .9745(.0018) & .7650(.0128) & .9482(.0067) \\
            PTN + \cite{Tanaka2018} in $D_s$ & .7576(.0118) & .9776(.0015) & .7743(.0029) & .9442(.0029) \\
            PTN + label distillation & .8073(.0043) & .9808(.0009) & .7941(.0060) & .9663(.0033) \\
            \hline \hline
        \end{tabular}
    \end{center}
\end{table}
\footnotetext{3 median results in perspective of validation accuracy}

\subsubsection{Photometric transformer networks}
We use 6 convolution layers for transformer networks and set $K$ to 10.
Each mammogram is resized to one-third of the original size.
We also use the instance normalization~\cite{Ulyanov2016} for each of the convolutions.

The standard deviation of image-level pixel mean values in the validation set is 0.6850.
Once PTN normalizes the validation images, the standard deviation is significantly reduced to 0.2249.
This verified that PTN suppresses inter-image intensity variations and make the image intensities more consistent.
Fig.~\ref{fig:qualitative_results} shows two normalization examples.

The upper part of Table~\ref{tab:algo} shows the results of normalization methods.
CLAHE~\cite{Pizer1987} is selected for representative of static normalization approach.
CLAHE improves the baseline, but PTN shows better performance.

\subsubsection{Label distillation}
For label distillation, we choose the best PTN model as our baseline.
The pre-trained parameters of the first two layers are fixed, and the rest is trained with a learning rate of 0.01.
We set $\alpha$ and $\gamma$ as 0.5 and 0.25, respectively.
The lower part of Table~\ref{tab:algo} shows the results.
The hard-labeling, which directly uses predictions of PTN as pseudo labels, shows the worst performance compared to the other methods.
\cite{Tanaka2018} is another approach, which uses soft pseudo labels.
To make \cite{Tanaka2018} fairly comaparable to our method, we fine-tune with $D_s$ at each epoch, before giving pseudo labels.
Overall, these two methods improve the baseline, however, the gains are not significant.
In contrast, our label distillation modthod yields clear performance gains compared to the baseline.

\subsection{Comparison with others}
\label{ssec:external_validation}
\begin{table}[t]
    \caption{Breast density estimation performance comparison between methods on the external test set.}
    \label{tab:external}
    \begin{center}
        \begin{tabular}{l|c|c}
            \hline \hline
            & Accuracy & dAUC \\ \hline
            LIBRA\footnotemark{} & - & .8877 \\
            \cite{Lehman2018} & .5860 & .9275 \\
            \cite{Wu2018} & .5419 & .8424 \\
            Our baseline & .4246 & .9185 \\
            Ours & .7257 & .9481 \\
            \hline \hline
        \end{tabular}
    \end{center}
\end{table}
\footnotetext{It produces the percent of density value directly, thus only dAUC is reported.}
We compare our models with LIBRA~\cite{Keller2012}, an open-source density predictor, and some other works \cite{Lehman2018,Wu2018} who have opened their model parameters in public.
In our baseline model, the PTN and the label distillation are not applied.
The results are shown in Table \ref{tab:external}.
For a fair comparison, our model is selected by the median value of accuracy score in the in-house validation set, among five trials of experiment.
Although our model is trained with the data consists of different race, our best model achieves the best performance with large margins in all metrics.

\section{Conclusion}

In this paper, we have proposed two methods for breast density problem: PTN and label distillation.
These two methods can resolve input and label issues in the breast density prediction task, respectively.
For further research, strict validation of dAUC metric how it is suitable for breast density tasks is needed.
Additionally, our approach should be looked in broad views, and applied for various medical imaging problems, since it
is not limited to a specific task.

\bibliographystyle{splncs04}
\bibliography{paper.bib}

\end{document}


\section{Baseline networks}
As described in the paper, we used ResNet-18 based architecture, but slight change applied to enlarge receptive fields.
